\documentclass[conference]{IEEEtran}
\usepackage{times}
\usepackage[numbers,sort&compress]{natbib}
\usepackage{multicol}
\usepackage[bookmarks=true,colorlinks=true,linkcolor=blue!60!black,citecolor=blue!60!black,urlcolor=blue!60!black]{hyperref}

\usepackage[dvipsnames,table]{xcolor}
\usepackage{amsmath,amssymb,amsthm}
\usepackage[ruled,lined,linesnumbered,commentsnumbered,longend]{algorithm2e}
\usepackage{graphicx}
\usepackage{booktabs}
\usepackage{multirow}
\let\labelindent\relax
\usepackage{enumitem}
\usepackage{siunitx}
\usepackage{xspace}
\usepackage{balance}

\newcommand{\sbf}[1]{{\small\textbf{#1}}\xspace}
\definecolor{myblue}{HTML}{0014A8}

\begin{document}

\title{Neural Fidelity Calibration for\\Informative Sim-to-Real Adaptation}


\author{\authorblockN{Youwei Yu \quad Lantao Liu}
\authorblockA{Luddy School of Informatics, Computing, and Engineering\\
Indiana University, Bloomington, IN 47408, USA\\
Email: \{youwyu, lantao\}@iu.edu}}

\maketitle

\begin{abstract}
Deep reinforcement learning (RL) can transfer agile robot skills from simulation to the real world, but closing the sim-to-real gap by domain randomization requires expert physics knowledge, and simulator calibration from real data typically targets a few hand-picked parameters. Even then, simulators miss real-world effects, and environments reconstructed from onboard perception inherit sensing errors that limit policy improvement.
We propose Neural Fidelity Calibration (NFC), which uses conditional flow matching to calibrate simulator physical parameters \emph{and} residual fidelity domains online during deployment. Residual fidelity captures sim-to-real dynamics discrepancies and perception-induced environment uncertainty. NFC operates in a latent space so that realistic environments can be sampled from the inferred posterior for real-to-sim policy fine-tuning. The framework is informative: the pretrained policy is fine-tuned only when an anomaly is detected, sequential NFC seeds its proposal from the simulation-pretrained estimator, and a hallucination policy modulates posterior uncertainty following optimism in the face of uncertainty.
NFC attains the most accurate posteriors among established simulation-based inference methods on five robots with up to \num{39} physical parameters and a $128\times128$ residual terrain. On a wheeled robot with a broken wheel axle deployed on snow, mud, and wet grass, NFC raises success on rough terrain from \SI{26}{\percent} to \SI{70}{\percent} and reduces orientation jerk by \SI{88}{\percent}. Code and videos will be released upon publication.
\end{abstract}

\IEEEpeerreviewmaketitle

\section{Introduction}
Zero-shot sim-to-real RL transfers agile policies to soft~\citep{tiboni23rfdroid}, wheeled~\citep{yu2024adaptive}, aerial~\citep{Geles-RSS-24}, and legged~\citep{Joonho2024wheelleg} robots, with domain randomization (DR) as the standard tool for robustness~\citep{tobin2017domain}. In practice, DR ranges are set by experts~\citep{ma2024dreureka}, environments are reconstructed by hand~\citep{escontrela2024learning}, and perturbations must be designed~\citep{Shi-RSS-24}; exhaustive coverage of parameters and environments is prohibitive.

The sim-to-real-to-sim paradigm trains in simulation and improves with real-world data~\citep{stachowicz2024lifelong}, keeping risky exploration in simulation. Within it, simulation-based inference (SBI)~\citep{papamak16efree} narrows DR by inferring simulator parameters from real trajectories, e.g., friction from a run on snow. Two obstacles remain in the field. First, SBI assumes a well-calibrated simulator, yet high-fidelity simulators omit real effects, and environments reconstructed from onboard sensors inherit perception errors. Second, kernel- and optimization-based estimators~\citep{ramos2019bayessim, sobanbabu2025sampling} target low-dimensional parameter sets and do not scale to the dimensionality encountered in deployment.

We propose Neural Fidelity Calibration (NFC), which uses conditional flow matching to calibrate simulator parameters and residual fidelity domains online while the robot operates. Our contributions are:
\begin{enumerate}[leftmargin=*, label=(\arabic*), itemsep=1pt]
\item \textbf{Simulator-to-reality shift.} We separate \emph{simulator calibration} (parameters aligning simulated with real dynamics) from \emph{residual fidelity}, which captures discrepancies remaining after correct calibration and errors in environments reconstructed from uncertain perception.
\item \textbf{Neural Fidelity Calibration.} NFC jointly infers both in a latent space with conditional flow matching, scaling to high-dimensional parameters and rich conditioning (state--action histories plus elevation or depth maps).
\item \textbf{Informative policy adaptation.} The policy is fine-tuned online only when an anomaly signals imminent failure; a hallucination policy, trained jointly with the control policy, modulates NFC uncertainty under optimism in the face of uncertainty; and all real data is reused offline.
\end{enumerate}
We evaluate NFC against established SBI methods on five simulated robots and on a real Jackal with a broken wheel axle across laboratory floors and wild terrain.

\section{Related Work}\label{sec:related}
\textbf{Robust sim-to-real transfer.} Zero-shot transfer has succeeded in locomotion~\citep{David24anymalparkour, zhuang2024humanoid}, navigation~\citep{Fabian24DTC, yu2024adaptive}, and manipulation~\citep{Handa23DeXtreme, Villasevil-RSS-24}, with DR bridging the gap. Adversarial training~\citep{Gleave2020Adversarial, Shi-RSS-24} and safe RL~\citep{Stachowicz-RSS-24, Hsu-RSS-21, hsuzen2022sim2lab2real, nguyen2024gameplay} harden policies further, but their guarantees hold relative to an assumed model and disturbance set, which an unmodeled fault can violate. This motivates continual real-world recalibration, which we trigger by anomaly detection.

\textbf{Adaptive domain randomization (ADR)} aligns simulator parameters with real executions~\citep{Heiden-RSS-21, tiboni23rfdroid, huang2023went}, mostly via likelihood-free SBI~\citep{papamak16efree}: BayesSim~\citep{ramos2019bayessim} uses random Fourier features, \citet{pmlr-v164-muratore22a} add sequential inference, and \citet{memmel2024asid} maximize Fisher information, while high-dimensional conditioning remains under-explored~\citep{greenberg2019automatic}. Evolution strategies~\citep{tsai21droid}, Bayesian optimization~\citep{ota21augsim}, and differentiable simulators~\citep{Heiden21neuralsim, Heiden22inf} are costly and still assume a simulator that can mirror reality; we drop that assumption by modeling the residual explicitly.

\textbf{Optimistic exploration.} Optimism in the face of uncertainty~\citep{brafman2002rmax, JMLR:v11:jaksch10a} underlies H-UCRL~\citep{curi2020efficient} and its extensions~\citep{sessa2022efficient, sukhija2023optimistic, bhardwaj24metarl}. Our hallucination policy borrows the principle but reshapes the spread of domains drawn from the NFC posterior rather than optimizing within a learned model's confidence set.

\section{Preliminaries and Problem Formulation}\label{sec:prelim}
\textbf{Simulation-based inference.} ADR infers simulator parameters $\boldsymbol{\psi}$ from a real trajectory $\boldsymbol{\tau}$ via $p(\boldsymbol{\psi}\vert\boldsymbol{\tau}) \propto p(\boldsymbol{\tau}\vert\boldsymbol{\psi})p(\boldsymbol{\psi})$. With a black-box simulator the likelihood is intractable, so likelihood-free methods~\citep{Sisson2007SMC, papamak16efree} approximate the posterior from simulated pairs. Their cost grows quickly with the dimension of $\boldsymbol{\psi}$, and existing estimators target a handful of parameters~\citep{ramos2019bayessim, sobanbabu2025sampling}, whereas our problems couple up to \num{39} physical parameters with a \num{16384}-dimensional residual environment, conditioned on trajectories with depth or elevation perception.

\textbf{Flow matching as posterior estimator.} We estimate the posterior with conditional flow matching~\citep{lipman2022flow}, which scales to high-dimensional targets and rich conditioning~\citep{wildberger2023flow}: a velocity field $\boldsymbol{v}_{\boldsymbol{\theta}}(\boldsymbol{\psi}(k),\boldsymbol{\tau},k)$ is regressed onto $\boldsymbol{\psi}(1)-\boldsymbol{\psi}(0)$ along the linear path from a Gaussian sample $\boldsymbol{\psi}(0)$ to a parameter sample $\boldsymbol{\psi}(1)$, and sampling integrates $\mathrm{d}\boldsymbol{\psi}/\mathrm{d}k = \boldsymbol{v}_{\boldsymbol{\theta}}$ over $k\in[0,1]$ with a few Euler steps, which makes online use practical. We write $\boldsymbol{q}_{\boldsymbol{\theta}}$ for the induced conditional density.

\textbf{Problem.} We model a contextual MDP~\citep{hallak2015cmdp} whose context $c$ comprises simulator parameters $\boldsymbol{\psi}$ and environment variables $\boldsymbol{e}$ (e.g., terrain geometry), and maximize return \emph{averaged over the posterior} of contexts consistent with the real execution:
\begin{equation}\label{eqn:problem-form}
\max_{\boldsymbol{\pi}} \; \mathbb{E}_{c\sim p(c\vert\boldsymbol{\tau}^{\mathrm{real}})}\,\mathbb{E}_{\boldsymbol{\tau}\sim P^{\boldsymbol{\pi}}_{c}}\left[R(\boldsymbol{\tau})\right].
\end{equation}
Calibrating to a single best-fit $\boldsymbol{\psi}^{*}$ instead overfits, since no single setting explains the data when the simulator is imperfect and real data is scarce. Because simulation budgets are finite~\citep{yu2024adaptive}, we target informative fine-tuning during deployment rather than exhaustive pretraining.

\begin{figure*}[t!]
\centering
\includegraphics[width=0.85\textwidth]{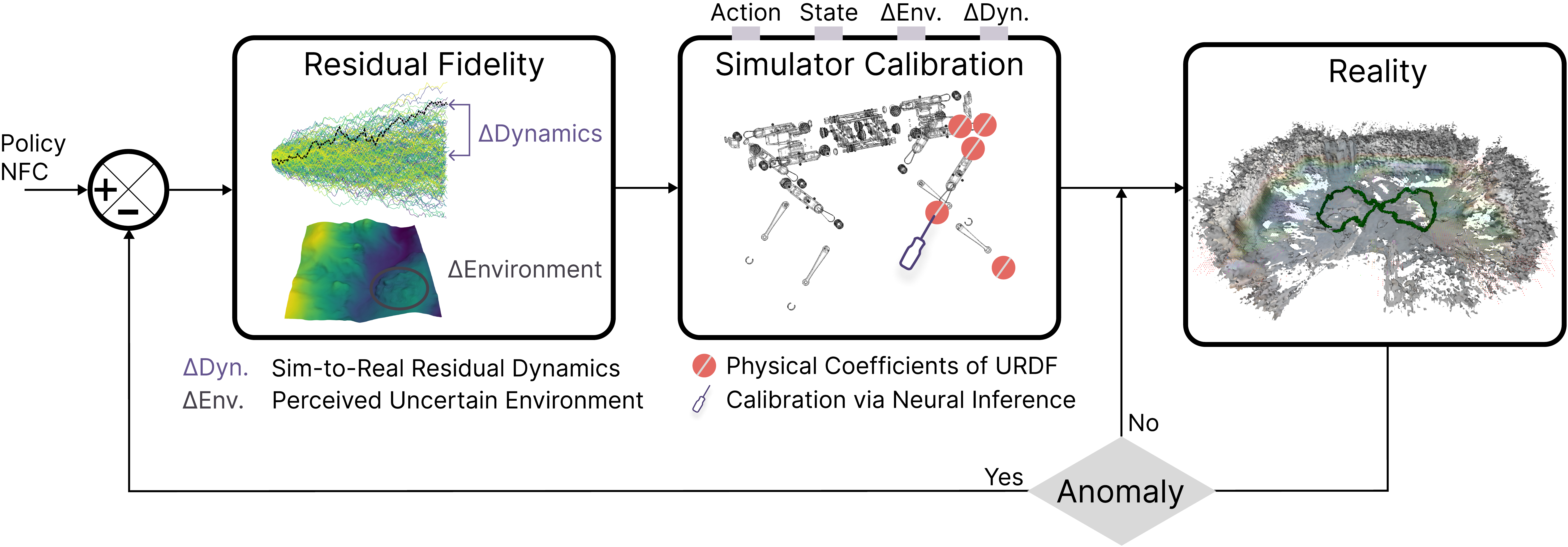}
\vspace*{-0.08in}
\caption{NFC-driven sim-to-real adaptation. (a)~\textbf{Residual fidelity}: colored trajectories are rollouts in the calibrated simulator, black is the real execution, and their difference is the residual dynamics; the circled region of the terrain map marks uncertain perception, where NFC samples the residual environment (ground-truth minus perceived elevation) to reconstruct terrain variants for fine-tuning. (b)~\textbf{Simulator calibration}: a conditional flow matching model infers physical parameters matching the real trajectory. (c)~\textbf{Informative adaptation}: policy and NFC are pretrained in simulation and fine-tuned only when an anomaly is detected.}
\label{fig:system}
\end{figure*}

\section{Neural Fidelity Calibration}\label{sec:method}
Fig.~\ref{fig:system} summarizes the framework.

\subsection{Calibration and Fidelity Shifts}
BayesSim-style inference explains the real trajectory only through the simulator's own parameterization, i.e., it assumes $p(\boldsymbol{\psi}\vert\boldsymbol{\tau}^{\mathrm{real}})\approx p(\boldsymbol{\psi}\vert\boldsymbol{\tau}^{\mathrm{sim}})$ for some $\boldsymbol{\psi}$. This fails when the simulator omits real effects and when it is calibrated against a terrain perceived with error~\cite{yu2025neural}. We therefore decompose real-world transitions as
\begin{equation}\label{eqn:sys-dynamics}
s_{t+1} = \underbrace{f_{\boldsymbol{\psi}}(s_t,a_t\vert\boldsymbol{e})}_{\mathrm{Calibration}} + \underbrace{g_{\boldsymbol{\phi}}(s_t,a_t\vert\boldsymbol{e})}_{\mathrm{Fidelity}} + \epsilon(s_t,a_t\vert\boldsymbol{e}),
\end{equation}
where $f_{\boldsymbol{\psi}}$ is the nominal simulator with parameters $\boldsymbol{\psi}$ (mass, friction, \dots), $g_{\boldsymbol{\phi}}$ models residual effects, and $\epsilon$ is noise. The residual fidelity $\boldsymbol{\phi}=\{\Delta s_{t+1},\Delta\boldsymbol{e}\}$ comprises the \emph{residual dynamics} $\Delta s_{t+1}$ (sim-vs-real state differences under identical conditions) and the \emph{residual environment} $\Delta\boldsymbol{e}$ (perceived vs.\ true environment). For a wheeled robot on rough terrain, $\boldsymbol{\psi}\in\mathbb{R}^{13}$ holds body and wheel masses, wheel damping, and friction; $\boldsymbol{\phi}$ holds residual dynamics over pose, velocity, and previous action ($\mathbb{R}^{15}$) and a residual elevation map $\Delta\boldsymbol{e}\in\mathbb{R}^{128\times128}$.

We assume $\boldsymbol{\psi}$, $\boldsymbol{\phi}$, and $\boldsymbol{\tau}$ have compact support with bounded densities, which keeps inferred parameters physically plausible (e.g., at most two broken wheels on a four-wheeled robot). NFC needs only these broad support bounds and a choice of how perturbations enter dynamics and observations; it requires no real-world anomaly labels or fault taxonomy. We then learn the joint posterior $p(\boldsymbol{\psi},\boldsymbol{\phi}\vert\boldsymbol{\tau})$ directly: a conditional density estimator $\boldsymbol{q}_{\boldsymbol{\theta}}$ is trained on simulator-generated parameter--trajectory pairs to maximize $\frac{1}{N}\sum_n \log \boldsymbol{q}_{\boldsymbol{\theta}}(\boldsymbol{\psi}_n,\boldsymbol{\phi}_n\vert\boldsymbol{\tau}_n)$, so the trajectory discrepancy enters implicitly through the simulated data-generating process rather than a hand-designed distance. For fine-tuning, $N$ posterior samples reconstruct $N$ environments $\boldsymbol{e}=\hat{\boldsymbol{e}}+\Delta\boldsymbol{e}$ that reset the $N$ parallel simulators.

\subsection{Sequential and Latent NFC}\label{sec:neural-dr}
\textbf{Sequential NFC.} At deployment, the simulation-pretrained estimator serves as proposal prior, $\tilde{p}(\boldsymbol{\psi},\boldsymbol{\phi}) = \boldsymbol{q}^{\mathrm{sim}}(\boldsymbol{\psi},\boldsymbol{\phi}\vert\boldsymbol{\tau}^{\mathrm{real}})$, concentrating online simulation on plausible parameters. Following~\citet{papamak16efree}, an estimator trained on proposal-generated data converges to $\boldsymbol{q}(\boldsymbol{\psi},\boldsymbol{\phi}\vert\boldsymbol{\tau}) \propto \frac{\tilde{p}(\boldsymbol{\psi},\boldsymbol{\phi})}{p(\boldsymbol{\psi},\boldsymbol{\phi})}\, p(\boldsymbol{\psi},\boldsymbol{\phi}\vert\boldsymbol{\tau})$, i.e., the true posterior reweighted by the proposal-to-prior ratio, which recovers the posterior when $\tilde{p}=p$. We use this density directly for fine-tuning; each online update trains on newly simulated trajectories only, warm-started from the current network, and is much faster than pretraining.

\textbf{Latent NFC.} Inferring $\Delta\boldsymbol{e}\in\mathbb{R}^{128\times128}$ in raw space is too costly for real-time improvement. An encoder $\mathcal{H}$---a causal temporal convolutional network (TCN)~\citep{BaiTCN2018} for the state--action history and an MLP for the fused perception map---compresses the execution data. In simulation both the noisy trajectory $\hat{\boldsymbol{\tau}}$ the robot would observe and the clean, privileged trajectory $\boldsymbol{\tau}$ are available, and their embedding difference defines the \emph{latent residual fidelity} $\boldsymbol{\phi}_z = \mathcal{H}(\boldsymbol{\tau}) - \mathcal{H}(\hat{\boldsymbol{\tau}})$. The estimator conditions on $\mathcal{H}(\hat{\boldsymbol{\tau}})$ and infers $\boldsymbol{q}(\boldsymbol{\psi},\boldsymbol{\phi}_z\vert\cdot)$; an MLP decoder $\mathcal{Z}$ maps $\boldsymbol{\phi}_z$ back to $\boldsymbol{\phi}$. Velocity field, encoder, and decoder are trained jointly on the negative log-density plus a reconstruction loss $\alpha\|\mathcal{Z}(\boldsymbol{\phi}_z)-\boldsymbol{\phi}\|_2^2$, alternating between updating $(\boldsymbol{q}_{\boldsymbol{\theta}},\mathcal{H})$ and $\mathcal{Z}$. The reconstruction term acts like privileged-information distillation in teacher--student transfer~\citep{he2024bridging}, and modeling the residual $\Delta\boldsymbol{e}$ rather than the full environment follows residual learning in image restoration~\citep{Yue2025Restoration}.

\subsection{Policy Optimization with Hallucinated Randomness}\label{sec:hallucinate}
Given NFC samples, the direct approach is DR with the posterior in place of hand-designed ranges. When the posterior is highly uncertain, however, sampling it directly can stall fine-tuning: very low friction makes the robot slide, very high friction makes it stick, and reward is scarce within a short online window. We introduce a hallucination policy $\boldsymbol{\pi}_{\mathfrak{h}}$ that never acts on the robot but controls how far each fine-tuning domain may deviate from the posterior center. Let $\boldsymbol{x}=[\boldsymbol{\psi},\boldsymbol{\phi}_z]\in\mathbb{R}^{D}$ be a joint latent sample and $\boldsymbol{\mu}$ the mean of the $N$ samples in the current round. Conditioned on the history embedding, $\boldsymbol{\pi}_{\mathfrak{h}}$ samples a gate $\boldsymbol{g}\in(0,1]^{D}$ per parallel environment, and each simulated domain is
\begin{equation}\label{eqn:hallucinate-domain}
\tilde{\boldsymbol{x}} = \boldsymbol{\mu} + \boldsymbol{g}\odot(\boldsymbol{x}-\boldsymbol{\mu}), \qquad \boldsymbol{x}\sim\boldsymbol{q}(\cdot\mid\hat{\boldsymbol{\tau}}),
\end{equation}
decoded by $\mathcal{Z}$. $\boldsymbol{g}=\boldsymbol{1}$ recovers the NFC posterior and $\boldsymbol{g}\to\boldsymbol{0}$ collapses to its mean; because the gate is per-coordinate, the agent can keep the variance of wheel friction while shrinking that of the residual-terrain latent. Control and hallucination policies are optimized jointly with PPO~\citep{schulman2017proximal},
\begin{equation}\label{eqn:final-policy-opt}
\max_{\boldsymbol{\pi},\boldsymbol{\pi}_{\mathfrak{h}}}\;
\mathbb{E}_{\boldsymbol{x}\sim\boldsymbol{q}(\cdot\mid\hat{\boldsymbol{\tau}})}\,
\mathbb{E}_{\boldsymbol{g}\sim\boldsymbol{\pi}_{\mathfrak{h}}(\cdot\mid\mathcal{H}(\hat{\boldsymbol{\tau}}))}\,
\mathbb{E}_{\boldsymbol{\tau}\sim P^{\boldsymbol{\pi}}_{\tilde{\boldsymbol{x}}}}\left[R(\boldsymbol{\tau})\right].
\end{equation}
Unlike model-based RL~\citep{curi2020efficient}, this treats the simulator as the model and learns its parameters rather than the dynamics.

\subsection{Informative Sim-to-Real Adaptation}\label{sec:adaptation}
Under nominal conditions the pretrained policy performs well and real data is kept for offline refinement; in failure-prone conditions we adapt online.

\textbf{Anomaly detection.} We compare a nominal reference window $\boldsymbol{\tau}^{\mathfrak{c}}$ with a target window $\boldsymbol{\tau}^{\mathfrak{t}}$~\citep{darban2025carla} as a classifier two-sample test~\citep{lopez2016revisiting}, instantiated with a Siamese network~\citep{koch2015siamese} whose encoder is the NFC TCN. The anomaly logit is $\ell_{\mathrm{A}} = \boldsymbol{w}^{\top}\vert\mathcal{H}(\boldsymbol{\tau}^{\mathfrak{c}})-\mathcal{H}(\boldsymbol{\tau}^{\mathfrak{t}})\vert + b$, trained with binary cross-entropy on simulated pairs only: nominal ($y=0$) when physical and residual parameters are retained, anomalous ($y=1$) when the target's parameters are redrawn from the prior. A window is flagged when $\ell_{\mathrm{A}}>\eta$, with $\eta$ the \num{99}th percentile on held-out nominal pairs.

\textbf{Deployment protocol (Alg.~\ref{alg:nfc}).} Policy, NFC, and detector are pretrained in simulation under DR. Online, each detected anomaly triggers one adaptation round: NFC infers the posterior from the latest $H$-step window, the hallucination policy gates its spread, gated samples reset $N$ parallel simulators, the policy is fine-tuned, and sequential NFC is updated with the gated samples as proposal. After the mission, the accumulated data refine NFC, detector, and policy.

\begin{algorithm}[t]
\caption{Informative Sim-to-Real Adaptation with NFC}
\label{alg:nfc}
\SetKwInOut{KwIn}{Input}
\KwIn{Policy $\boldsymbol{\pi}$, NFC $\boldsymbol{q}$ with $\mathcal{H},\mathcal{Z}$, anomaly detector, hallucination policy $\boldsymbol{\pi}_{\mathfrak{h}}$, all pretrained in simulation}
\While{mission not complete}{
    Execute window $\boldsymbol{\tau}_t$ of $H$ steps\;
    $\ell_{\mathrm{A}} \gets \boldsymbol{w}^{\top}\vert\mathcal{H}(\boldsymbol{\tau}^{\mathfrak{c}})-\mathcal{H}(\boldsymbol{\tau}_t)\vert + b$ \tcp*{anomaly test}
    \If{$\ell_{\mathrm{A}} > \eta$}{
        $\{\boldsymbol{x}_n\}_{n=1}^{N}\sim\boldsymbol{q}(\cdot\vert\mathcal{H}(\boldsymbol{\tau}_t))$, $\boldsymbol{\mu}\gets\frac{1}{N}\sum_n\boldsymbol{x}_n$ \tcp*{NFC}
        $\boldsymbol{g}_n\sim\boldsymbol{\pi}_{\mathfrak{h}}(\cdot\mid\mathcal{H}(\boldsymbol{\tau}_t))$, $\tilde{\boldsymbol{x}}_n\gets\boldsymbol{\mu}+\boldsymbol{g}_n\odot(\boldsymbol{x}_n-\boldsymbol{\mu})$ \tcp*{Eq.~\eqref{eqn:hallucinate-domain}}
        $\tilde{\boldsymbol{\phi}}_n\gets\mathcal{Z}(\tilde{\boldsymbol{\phi}}_{z,n})$, $\boldsymbol{e}_n\gets\hat{\boldsymbol{e}}+\Delta\boldsymbol{e}_n$ \tcp*{reset $N$ simulators}
        $\boldsymbol{\pi},\boldsymbol{\pi}_{\mathfrak{h}}\gets$ \texttt{Policy-Opt} on $\{\tilde{\boldsymbol{\psi}}_n,\tilde{\boldsymbol{\phi}}_n\}$, Eq.~\eqref{eqn:final-policy-opt}\;
        $\boldsymbol{q}\gets$ train on $\{(\tilde{\boldsymbol{x}}_n,\boldsymbol{\tau}^{\mathrm{sim}}_n)\}$ \tcp*{sequential NFC}
    }
    $\mathcal{T}\gets\mathcal{T}\cup\boldsymbol{\tau}_t$\;
}
Refine $\boldsymbol{\pi}$, $\boldsymbol{q}$, and detector offline with $\mathcal{T}$\;
\end{algorithm}

\section{Simulation Experiments}\label{sec:sim-exp}
\textbf{Setup.} We evaluate \texttt{Ant}, \texttt{Quadruped} (Unitree Go1), \texttt{Humanoid} (Unitree H1), \texttt{Quadcopter} (Ingenuity), and \texttt{Jackal} (ClearPath), with $\vert\boldsymbol{\psi}\vert = 17, 25, 39, 9, 13$ physical parameters (masses, stiffness or rotor gains, damping, friction) and residual fidelity over proprioceptive state plus a $128\times128$ residual elevation map (a residual depth map for \texttt{Quadcopter}). Policies and NFC are pretrained in Genesis~\citep{Genesis} and deployed in MuJoCo~\citep{todorov2012mujoco}, whose different contact and integration models plus added sensing and actuation noise provide the gap, in \emph{Flat} (obstacle-free) and \emph{Rough} (randomized elevation maps~\citep{yu2024adaptive} or 3D obstacle fields~\citep{kulkarni2023aerialgymisaac}) environments. Policies are PPO with \num{4096} parallel environments; NFC pretraining uses \num{1e6} windows of $H=50$ steps at \SI{25}{\hertz}. Each online fine-tuning round takes \SI{0.2}{\second} (\SI{5}{\hertz} policy improvement); online NFC training (\SI{1.75}{\second}) runs in parallel with the \SI{2}{\second} execution window.

\textbf{Baselines.} With our flow matching (FM) head we compare feature extractors \sbf{MLP} (also our ablation without latent reconstruction), \sbf{QMC-RFF}~\citep{ramos2019bayessim}, and \sbf{QMC-FKS}~\citep{johannes2024slice}; against the FM head we compare mixture density networks \sbf{MDN-RFF} (BayesSim) and \sbf{MDN-TCN}, and \sbf{CMA-ES}~\citep{hansen2016cma, sobanbabu2025sampling}, which minimizes trajectory MSE under identical actions.

\begin{table*}[t!]
\small
\centering
\setlength{\tabcolsep}{2.4pt}
\caption{Posterior-inference accuracy (average log-density of ground-truth parameters, $\uparrow$) and anomaly detection (TPR $\uparrow$ / FPR $\downarrow$ at a threshold calibrated to \SI{99}{\percent} specificity in Genesis). -- denotes not applicable.}
\vspace*{-0.15in}
\label{tab:anomaly-detect}
\begin{tabular*}{\textwidth}{@{\extracolsep{\fill}} ll rrrrrrrrrr rrrr @{}}
\toprule
& & \multicolumn{10}{c}{\textbf{Average Log Posterior Density} ($\uparrow$)} & \multicolumn{4}{c}{\textbf{Anomaly Detection}} \\
\cmidrule(lr){3-12} \cmidrule(lr){13-16}
& & \multicolumn{2}{c}{\texttt{Ant}} & \multicolumn{2}{c}{\texttt{Quadruped}} & \multicolumn{2}{c}{\texttt{Humanoid}} & \multicolumn{2}{c}{\texttt{Quadcopter}} & \multicolumn{2}{c}{\texttt{Jackal}} & \multicolumn{2}{c}{TPR ($\uparrow$)} & \multicolumn{2}{c}{FPR ($\downarrow$)} \\
\cmidrule(lr){3-4} \cmidrule(lr){5-6} \cmidrule(lr){7-8} \cmidrule(lr){9-10} \cmidrule(lr){11-12} \cmidrule(lr){13-14} \cmidrule(lr){15-16}
\textbf{Inference} & \textbf{Feature} & \emph{Flat} & \emph{Rough} & \emph{Flat} & \emph{Rough} & \emph{Flat} & \emph{Rough} & \emph{Flat} & \emph{Rough} & \emph{Flat} & \emph{Rough} & \emph{Flat} & \emph{Rough} & \emph{Flat} & \emph{Rough} \\
\midrule
MDN    & QMC-RFF & $-4.88$ & $-0.59$ & $-2.09$ & $-2.30$ & $-10.52$ & $-13.27$ & $-2.26$ & $-2.47$ & $0.64$ & $-1.20$ & $0.70$ & $0.51$ & $0.01$ & $0.01$ \\
MDN    & TCN     & $-2.60$ & $-3.10$ & $-1.64$ & $-3.11$ & $-2.52$  & $-3.42$  & $-1.79$ & $-2.56$ & $0.22$ & $-1.17$ & $\mathbf{0.97}$ & $\mathbf{0.80}$ & $0.01$ & $0.07$ \\
FM    & MLP     & $-1.00$ & $-1.09$ & $-1.14$ & $-1.34$ & $-1.63$  & $-2.93$  & $-1.20$ & $-1.86$ & $0.26$ & $-1.10$ & $0.85$ & $0.65$ & $0.01$ & $0.01$ \\
FM    & QMC-RFF & $-2.79$ & $-3.03$ & $-2.17$ & $-2.48$ & $-2.61$  & $-3.33$  & $-3.00$ & $-3.64$ & $0.06$ & $-1.29$ & $0.70$ & $0.51$ & $0.01$ & $0.01$ \\
FM    & QMC-FKS & $-2.63$ & $-3.05$ & $-2.00$ & $-1.91$ & $-2.41$  & $-3.37$  & $-2.42$ & $-2.73$ & $0.24$ & $-1.16$ & $0.33$ & $0.32$ & $0.01$ & $0.00$ \\
CMA-ES & N/A     & $-5.67$ & $-5.91$ & $-6.25$ & $-6.70$ & $-13.16$ & $-10.54$ & $-2.65$ & $-5.04$ & $-0.31$ & $-1.35$ & -- & -- & -- & -- \\
\midrule
FM    & TCN (Ours) & $\mathbf{0.57}$ & $\mathbf{0.46}$ & $\mathbf{-0.42}$ & $\mathbf{-0.46}$ & $\mathbf{-1.03}$ & $\mathbf{-2.64}$ & $\mathbf{-0.90}$ & $\mathbf{-1.50}$ & $\mathbf{1.21}$ & $\mathbf{0.69}$ & $\mathbf{0.97}$ & $\mathbf{0.80}$ & $0.01$ & $0.07$ \\
\bottomrule
\end{tabular*}
\end{table*}

\begin{figure}[!t]
\centering
\includegraphics[width=0.46\textwidth]{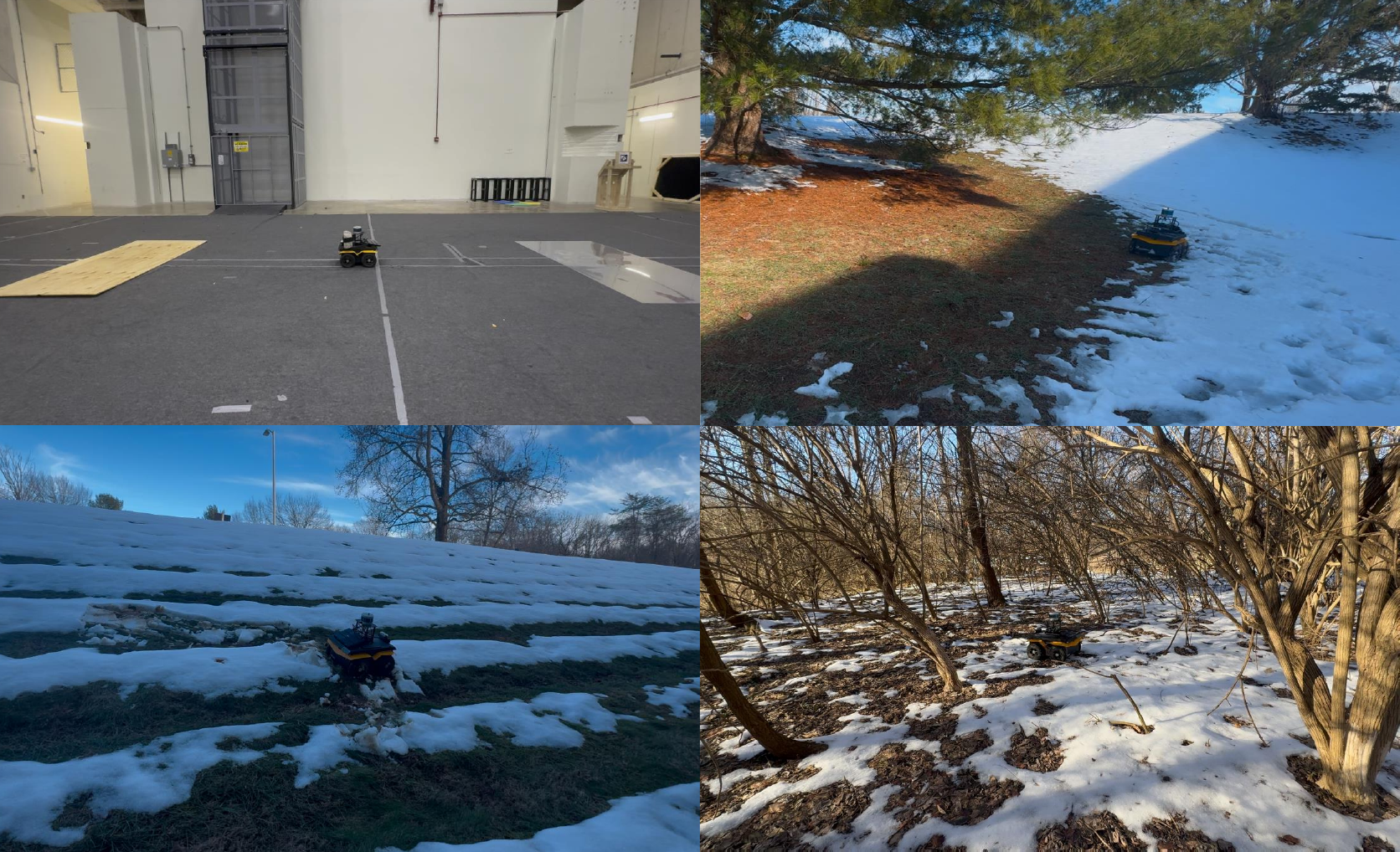}
\vspace*{-0.1in}
\caption{Real-world scenarios for a Jackal with a broken front-left wheel axle. Laboratory (\emph{Flat}) conditions vary surface material and a loosely attached rock; outdoor (\emph{Rough}) conditions combine snow, wet grass, mud, and elevation change.}
\label{fig:real-world-envs}
\end{figure}

\textbf{Anomaly detection.} Detectors are trained only on Genesis pairs and evaluated on \num{10000} nominal and \num{10000} anomalous MuJoCo windows per robot--terrain setting, with anomalous parameters in the tails of the prior (Table~\ref{tab:anomaly-detect}, right). With the false-alarm rate calibrated, methods separate on TPR: the TCN detector misses a few percent of windows on \emph{Flat} and one in five on \emph{Rough}, kernel features over a third. In deployment it flags \SIrange{96}{100}{\percent} of anomalous episodes.

\textbf{NFC accuracy.} Measured by the average log-density of ground-truth parameters under the predicted posterior~\citep{ramos2019bayessim}, NFC is best in all ten settings (Table~\ref{tab:anomaly-detect}, left). Accuracy drops from \emph{Flat} to \emph{Rough}, where perception enters the conditioning, and as $\vert\boldsymbol{\psi}\vert$ grows. FM-MLP is the strongest baseline, so the flow matching head already improves over MDN; the TCN encoder adds the remaining margin, and fixed random bases (RFF, FKS) trail because they cannot adapt to high-dimensional conditioning. CMA-ES converges offline given \num{100} rather than \num{10} simulation iterations, but each iteration re-runs the simulator, which is incompatible with online adaptation.

\textbf{Policy improvement.} We compare \sbf{PPO w/ NFC} with ablations \sbf{PPO w/o $\Delta$Fidelity} (physical calibration only) and \sbf{PPO w/o NFC} (fine-tuned directly on execution data), with \sbf{PPO w/ MDN-RFF} and \sbf{PPO w/ CMA-ES}, and with the model-based \sbf{TD-MPC2}~\citep{hansen2024tdmpc}, which fine-tunes its dynamics model and policy from execution data alone. In normalized return, PPO w/ NFC is best in all ten robot--terrain settings (e.g., \texttt{Jackal} \emph{Rough}: \num{0.82} vs.\ \num{0.69} w/o $\Delta$Fidelity, \num{0.20} w/o NFC, \num{0.52} MDN-RFF, \num{0.50} CMA-ES, \num{0.41} TD-MPC2). Removing residual fidelity lowers return in all ten settings, with the largest drop on \texttt{Quadruped} (\num{0.34}) and the smallest on \texttt{Humanoid} (\num{0.03}), so its contribution depends on embodiment rather than robot complexity; disabling NFC entirely causes the largest loss. TD-MPC2 beats PPO w/o NFC in nine of ten settings but never PPO w/ NFC, and its perception-aware dynamics model runs at only \SI{6}{\hertz}.

\begin{figure*}[t!]
\centering
\includegraphics[width=0.84\textwidth]{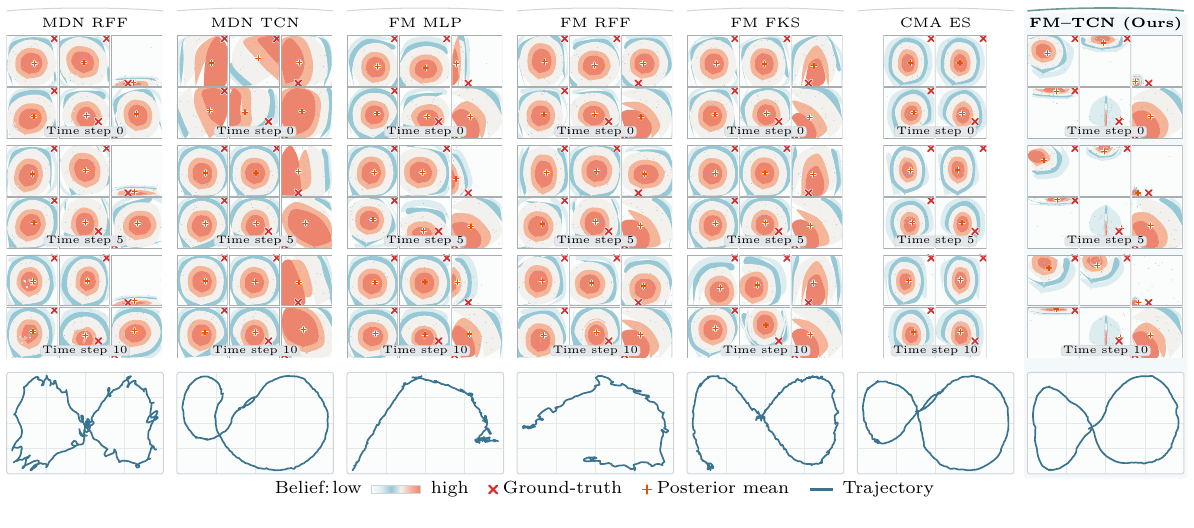}
\vspace*{-0.25in}
\caption{NFC on real snowy terrain with a broken front-left wheel axle. For each history window, each method block shows a $3\times2$ grid: the first two columns are joint posteriors over wheel damping ($x$) and friction ($y$) for the four wheels (top: rear-right, rear-left; bottom: front-right, front-left); the third column shows the inferred residual dynamics ($\Delta x,\Delta y$; top) and residual terrain height ($\Delta z$; bottom). NFC singles out the front-left wheel with near-zero damping; competing methods trained on the same data yield diffuse posteriors.}
\label{fig:real-world-posterior}
\end{figure*}

\begin{table*}[t]
\small
\centering
\setlength{\tabcolsep}{5pt}
\caption{Real-world policy improvement for a Jackal with a broken front-left wheel axle across four \emph{Flat} (laboratory) and three \emph{Rough} (wild terrain) scenarios, \num{9} runs each.}
\vspace*{-0.15in}
\label{tab:real-world-policy-improve}
\begin{tabular}{ll ccccccc}
\toprule
& & \begin{tabular}[c]{@{}c@{}}PPO\\ w/ NFC\end{tabular} & \begin{tabular}[c]{@{}c@{}}PPO\\ w/o $\Delta$Fidelity\end{tabular} & \begin{tabular}[c]{@{}c@{}}PPO\\ w/o NFC\end{tabular} & \begin{tabular}[c]{@{}c@{}}PPO\\ w/ MDN-RFF\end{tabular} & \begin{tabular}[c]{@{}c@{}}PPO\\ w/ CMA-ES\end{tabular} & TD-MPC2 & Falco \\
\midrule
\multirow{2}{*}{\begin{tabular}[c]{@{}l@{}}Success Rate\\ (\%) ($\uparrow$)\end{tabular}}
  & \emph{Flat}  & $\mathbf{100}$ & $\mathbf{100}$ & $\mathbf{100}$ & $\mathbf{100}$ & $\mathbf{100}$ & $\mathbf{100}$ & $\mathbf{100}$ \\
  & \emph{Rough} & $\mathbf{70}$  & $52$  & $26$  & $48$  & $44$  & $41$  & $30$ \\
\multirow{2}{*}{Traj.\ Ratio ($\downarrow$)}
  & \emph{Flat}  & $1.80$ & $1.80$ & $1.90$ & $1.53$ & $1.99$ & $\mathbf{1.50}$ & $1.60$ \\
  & \emph{Rough} & $\mathbf{2.10}$ & $3.20$ & $3.70$ & $2.40$ & $2.35$ & $2.70$ & $3.60$ \\
\multirow{2}{*}{\begin{tabular}[c]{@{}l@{}}Orien.\ Jerk ($\downarrow$)\\ (\si{\radian\per\second\cubed})\end{tabular}}
  & \emph{Flat}  & $\mathbf{2.98}$ & $3.50$  & $14.18$ & $6.14$ & $7.96$  & $6.95$  & $41.90$ \\
  & \emph{Rough} & $\mathbf{3.19}$ & $8.49$  & $26.99$ & $9.81$ & $10.38$ & $8.98$  & $41.93$ \\
\multirow{2}{*}{\begin{tabular}[c]{@{}l@{}}Pos.\ Jerk ($\downarrow$)\\ (\si{\meter\per\second\cubed})\end{tabular}}
  & \emph{Flat}  & $\mathbf{4.36}$ & $9.32$  & $8.41$  & $6.92$ & $7.79$  & $5.27$  & $6.95$ \\
  & \emph{Rough} & $\mathbf{6.26}$ & $12.46$ & $12.42$ & $9.39$ & $11.44$ & $18.46$ & $8.87$ \\
\bottomrule
\end{tabular}
\end{table*}

\section{Real-World Experiments}\label{sec:real-world}
\textbf{Setup.} The ClearPath Jackal carries a RealSense D435i, a Velodyne LiDAR, and an IMU; depth and LiDAR-inertial odometry are fused onboard into elevation maps~\citep{RAL18-elemap,fasterlio,yu2023mimu}, and policy optimization runs on a laptop GPU. All scenarios use a \emph{broken front-left wheel axle} (Fig.~\ref{fig:real-world-envs}). \emph{Flat} comprises four laboratory challenges (standard floor; floor with a loosely attached rock on the robot; mixed materials; mixed materials with the rock) and \emph{Rough} three wild-terrain challenges (wet grass and snow; elevated terrain with wet grass and snow; mud and snow in a forest). The task is a figure-eight course of \SI{3}{\meter} diameter followed as waypoints; a lap succeeds when every waypoint is reached within \SI{0.25}{\meter} without rollover or human intervention. Each controller is run \num{9} times per scenario; the detector flagged all \num{2160} anomalous trajectories.

\textbf{Baselines and metrics.} We compare NFC against FM-MLP, FM-RFF, FM-FKS, and MDN-RFF for calibration, and PPO w/ NFC against its ablations, PPO w/ MDN-RFF, PPO w/ CMA-ES, TD-MPC2, and \sbf{Falco}~\citep{Zhang2020Falco}, a high-rate motion-primitive planner with tuned \SI{200}{\hertz} PID control and access to elevation maps~\citep{cao2023granularity, Yang-RSS-23}. Following~\citep{siva2021enhancing, yu2024adaptive} we report success rate, trajectory ratio (traveled over target path length; \num{1.0} is ideal), orientation jerk, and position jerk.

\textbf{Calibration results.} Fig.~\ref{fig:real-world-posterior} shows calibration and residual fidelity on snow. The calibrated damping of the front-left wheel averages \num{0}, consistent with an undriven, free-spinning wheel, while friction on all four wheels averages \num{0.4}, reflecting the snow. Competing methods trained on the same data produce diffuse posteriors that do not single out the broken wheel. For the residual environment, the onboard map is noisy and incomplete because of complex geometry and materials, and NFC samples plausible, diverse terrain reconstructions (Fig.~\ref{fig:system}a). Ground truth for residual fidelity is unavailable in the real world, so its effect is measured through policy improvement.

\textbf{Policy improvement.} Table~\ref{tab:real-world-policy-improve} summarizes the results. First, NFC is essential: PPO fine-tuned only on real data (PPO w/o NFC) succeeds on \emph{Flat} but oscillates with high jerk, handling the anomaly in outcome but not in motion quality, and on \emph{Rough} falls into repetitive back-and-forth motion. Second, residual fidelity matters: PPO w/o $\Delta$Fidelity, which injects the perceived map directly into simulation, improves over PPO w/o NFC (\SI{52}{\percent} vs.\ \SI{26}{\percent} \emph{Rough} success) but stays well below the full method; the gap is attributable to the residual dynamics and perception uncertainty that only $\Delta$Fidelity models. Third, Falco attains low position jerk from its \SI{200}{\hertz} loop but the highest orientation jerk because its primitives ignore the asymmetric traction of the broken axle, and TD-MPC2 is sample-efficient but computationally heavy once perception enters state propagation. Overall, NFC raises \emph{Rough} success from \SI{26}{\percent} to \SI{70}{\percent} while reducing orientation jerk by \SI{88}{\percent}, handling a broken axle and unstructured terrain together.

\section{Limitations and Conclusion}\label{sec:conclusion}
Detection contrasts a reference and a target window, so slowly developing changes produce little signal and are absorbed only offline. Parameters and residuals need not be uniquely identifiable from a short trajectory, so inferred values should be read together with the downstream policy result. Fine-tuning updates persist across rounds, so returning to an earlier regime does not by itself restore earlier behavior; regime-indexed policies selected by the NFC posterior, and pairing the anomaly trigger with a recovery policy, are natural extensions.

NFC jointly calibrates simulator physics and residual fidelity---residual dynamics and residual environment---for informative online fine-tuning. With sequential latent NFC and optimistic hallucination, it delivers accurate posteriors at real-time cost and, on real rough terrain with a broken axle, lifts success from \SI{52}{\percent} to \SI{70}{\percent} over calibration alone.

\balance
\bibliographystyle{plainnat}
\bibliography{references}

\end{document}